# Active inference, Bayesian optimal design, and expected utility

Noor Sajid[1*], Lancelot Da Costa[1,2], Thomas Parr[1] and Karl Friston[1]
{noor.sajid.18, thomas.parr.12, k.friston}@ucl.ac.uk, l.da-costa@imperial.ac.uk
[1]Wellcome Centre for Human Neuroimaging, University College London, London, UK
[2]Department of Mathematics, Imperial College London, London, UK
* Correspondence to

**Abstract** Active inference, a corollary of the free energy principle, is a formal way of describing the behavior of certain kinds of random dynamical systems—that have the appearance of sentience. In this chapter, we describe how active inference combines Bayesian decision theory and optimal Bayesian design principles under a single imperative to minimize expected free energy. It is this aspect of active inference that allows for the natural emergence of information-seeking behavior. When removing prior outcomes preferences from expected free energy, active inference reduces to optimal Bayesian design, i.e., information gain maximization. Conversely, active inference reduces to Bayesian decision theory in the absence of ambiguity and relative risk, i.e., expected utility maximization. Using these limiting cases, we illustrate how behaviors differ when agents select actions that optimize expected utility, expected information gain, and expected free energy. Our T-maze simulations show optimizing expected free energy produces goal-directed information-seeking behavior while optimizing expected utility induces purely exploitive behavior—and maximizing information gain engenders intrinsically motivated behavior.

## 1.0 Introduction

Humans contend with conflicting objectives when operating in capricious, non-stationary environments, including maximizing epistemic value or minimizing expected cost (Laureiro-Martínez, Brusoni, & Zollo, 2010; Schulz & Gershman, 2019; Schwartenbeck et al., 2019). A widely studied proposition, for understanding how the balance between these distinct imperatives is maintained, is active inference (K. Friston, FitzGerald, Rigoli, Schwartenbeck, & Pezzulo, 2017; K. J. Friston, M. Lin, et al., 2017) – bringing together perception and action under a single objective of minimizing free energy across time (Da Costa, Parr, et al., 2020; Kaplan & Friston, 2018). Briefly, active inference is a formal way of describing the behavior of self-organizing (random dynamical) systems that interface with the external world, such as humans in their environment, with latent representations that maintain a consistent form (i.e., a particular steady-state) over time.

Active inference stipulates that by minimizing free energy (i.e., evidence upper bound) across time, agents will maintain sensed outcomes within a certain hospitable range (K. Friston, 2019; K. Friston et al., 2014). For this, agents optimize two complementary free energy functionals across time: variational and expected free energy[1]. Variational free energy measures the fit between an internal generative model and observed outcomes, while expected free energy scores each possible action trajectory in terms of its ability to reach a range of preferred latent states. Expected free energy equips the agent with a formal way to assess different hypothesis about the types of behavior that can be pursued, that guarantee realization of the agents' preferences over states and model parameters; c.f., planning as inference (Attias, 2003; Botvinick & Toussaint, 2012; Kaplan & Friston, 2018). The imperative to minimize expected free energy (or maximize expected model evidence), subsumes several important objectives that are prevalent in the psychological, economics, and engineering literature: e.g., KL control (Todorov, 2008; van den Broek, Wiegerinck, & Kappen, 2010), expected utility theory (Fleming & Sheu, 2002; Kahneman & Tversky, 1979), etc. These special cases are revealed by removing particular sources of uncertainty from the problem setting.

In this chapter, we show how the expected free energy combines both: Bayesian decision theory and optimal Bayesian design (Da Costa, Parr, et al., 2020; K. Friston, Da Costa, Hafner, Hesp, & Parr, 2020). Specifically, by removing prior preferences about outcomes from the expected free energy, active inference reduces to optimal Bayesian design (Chaloner & Verdinelli, 1995; Lindley, 1956; Pukelsheim, 2006; Stone, 1959). That is, it is only concerned with resolving uncertainty about the causes that generated particular outcomes. In contrast, by removing ambiguity and (posterior) predictive entropy from the expected free energy objective, active inference reduces to Bayesian decision theory (Harsanyi, 1978; Savage, 1972); namely, the maximization of some utility function, expected under predictive posterior beliefs about the consequences of an action. Hence, Bayesian decision theory and optimal Bayesian design can be combined to equip agents with a rich trade-off between information-seeking and goal-directed behaviors.

Effectively, expected free energy minimization introduces a Bayes-optimal arbitration between intrinsic, information-seeking, and extrinsic goal-seeking behavior (Karl J. Friston et al., 2015; Sajid, Ball, Parr, & Friston, 2021; Schwartenbeck et al., 2019). Here, information-seeking or uncertainty resolving behavior comes in two flavors. The first is uncertainty about beliefs about the state of the world and how they unfold (Parr & Friston, 2019a): agents will actively sample outcomes that have an unambiguous (i.e., low conditional entropy) relationship to latent states. Second, this kind of behavior resolves uncertainties about beliefs over particular model parameters (Schwartenbeck et al., 2019). Therefore, agents will expose themselves to observations that enable learning of the probabilistic structure of unknown—and unexplored—contingencies (K. J. Friston, M. Lin, et al., 2017; Schmidhuber, 2006). Consequently, expected free energy minimization entails information-seeking

[1] Both free energies are closely related: systems at steady-state that minimize variational free energy also minimize expected free energy (Parr, Da Costa, & Friston, 2020)

behavior that equips active inference agents with a deterministic way to explore unknown states—responding optimally to epistemic affordances.

We review these aspects of active inference and show that the minimization of expected free energy subsumes Bayesian decision theory and optimal Bayesian design principles as special cases. First, we briefly introduce active inference. This sets the scene to derive optimal Bayesian design principles and Bayesian decision theory as limiting cases of active inference. We then evaluate differences in behavior simulated under active inference; namely, optimal Bayesian design (i.e., by removing prior preferences about outcomes) and Bayesian decision theory (i.e., by removing ambiguity and relative risk)—in a T-maze setting. For simplicity, we focus on uncertainty stemming about states of the world, noting the same principles apply to lawful contingencies, parameterized by a generative model. The ensuing simulations show that goal-directed, information-seeking behavior is a direct consequence of minimizing expected free energy, under active inference (Millidge, Tschantz, & Buckley, 2020; Parr & Friston, 2019b). That is, agents selectively sample options that are associated with the highest information gain, i.e., the most informative, when engaging with their environment. Conversely, expected utility maximizing agents exhibit sub-optimal behavior by failing to sample—and exploit—epistemic or informative cues (Tschantz, Seth, & Buckley, 2020). We conclude with a brief discussion about the relevance of active inference as a way to characterize and quantify information-seeking behavior, in relation to goal-seeking.

## 2.0 Active Inference

Active inference describes how goal-directed, information-seeking agents navigate their environments (Da Costa, Parr, et al., 2020; K. Friston et al., 2017). For this, it stipulates three essential components: 1) a generative model of the agent's environment (Eq.1), 2) fitting the model to (sampled) observations to reduce surprise (i.e., variational free energy—Eq.2) and 3) selecting actions that minimize uncertainty (i.e., expected free energy—Eq.3).

This is formalized as a partially observable Markov decision process (POMDP), with the following random variables:

- $s \in S$; where $s$ is a particular latent state and $S$ a finite set of all possible latent states,
- $o \in O$; where $o$ is a particular observation and $O$ a finite set of all possible observations,
- $\pi \in \Pi$; where $\pi = \{a_1, a_2, ..., a_T\}$ is a policy (i.e., action trajectory) and $\Pi$ a finite set of all possible policies up to a given time horizon $T \in N^+$, and
- $T = \{1,..t,..\tau,T\}$; a finite set which stands for discrete time; $t$ and $\tau$ are some current and future time, respectively.

From this, we define the agent's generative model as the following probability distribution—omitting model parameters for simplicity:

$$P(o_{1:T}, s_{1:T}, \pi) = \underbrace{P(s_1)P(\pi)}_{\text{Priors}} \left( \prod_{t=1}^{T} \underbrace{P(o_t \mid s_t)}_{\text{Likelihood}} \right) \left( \prod_{t=2}^{T} \underbrace{P(s_t \mid s_{t-1}, \pi)}_{\text{Transition}} \right) \tag{1}$$

Accordingly, the agent minimizes surprise about observations (i.e., fits a model to observations) i.e., $-\log P(o_{1:t})$, through optimization of the following objective:

$$-\log P(o_{1:t}) \leq \mathrm{E}_{Q(s_{1:T}, \pi)}[\log Q(s_{1:T}, \pi) - \log P(o_{1:t}, s_{1:T}, \pi)] \tag{2}$$

where $Q(s_{1:T}, \pi)$ is the variational distribution over $s$ and $\pi$. Eq.2 undergirds the imperative to update beliefs about latent states to align them with observed outcomes. The inequality is derived using Jensen's inequality. The right-hand side is commonly referred to as variational free energy ($F$) (Karl J. Friston, 2010; Karl J Friston, Daunizeau, Kilner, & Kiebel, 2010) or the evidence lower bound in the variational inference literature (Beal, 2003; Blei, Kucukelbir, & McAuliffe, 2017). Furthermore, uncertainty about anticipated observations is reduced by selecting policies that *a priori* minimize expected free energy ($G$); $\log P(o_\tau)$ where $\tau \geq t$ (Parr & Friston, 2019b):

$$G(\pi, \tau) = \mathrm{E}_{P(o_\tau | s_\tau) Q(s_\tau | \pi)}[\log Q(s_\tau \mid \pi) - \log P(o_\tau, s_\tau)] \tag{3}$$

Note the resemblance to the terms in Eq.2, which can be transformed into Eq.3 by supplementing the expectation under the approximate posterior with the likelihood, resulting in the following predictive distribution: $\tilde{Q} = P(o_\tau \mid s_\tau) Q(s_\tau \mid \pi)$. This treats planning as inference (Attias, 2003; Botvinick & Toussaint, 2012); where we can evaluate plausible policies before outcomes have been observed. Additionally, we condition upon the policy in the approximate posterior and omit the policy from the generative model. Consequently, expected free energy is the main construct of interest because it determines the behavior of an agent when planning or selecting its course of action. We reserve description of its main features and relationship to other Bayesian objectives for the next section.

In active inference, decision-making entails the selection of the most likely action from the following distribution:

$$P(\pi) = \sigma(-G(\pi)) = \sigma\left( -\sum_{\tau > t} G(\pi, \tau) \right) \tag{4}$$

where $\sigma(.)$ is the softmax function and $G(\pi)$ is the expected free energy of a policy. The action associated with the sampled policy at the next timestep is then selected.

Using these components, expectations about latent states and policies can be optimized through inference and model parameters optimized through learning. This involves converging on the solution using gradient descent on free energy, $F$, which offers a biophysically plausible account of inference and learning (Da Costa, Parr, et al., 2020; K. Friston et al., 2017; Parr, Markovic, Kiebel, & Friston, 2019), and furnishing the expected free energy of various actions (Çatal, Wauthier, Verbelen, De Boom, & Dhoedt, 2020; Fountas, Sajid, Mediano, & Friston, 2020; van der Himst & Lanillos, 2020).

## 3.0 Expected free energy decomposed

Expected free energy, a central quantity within active inference, is (loosely speaking) the free energy functional of future trajectories. It evaluates the goodness of plausible policies determined by outcomes that have yet to be observed, i.e., anticipated future outcomes (Parr & Friston, 2019b). Accordingly, its minimization allows the agent to influence the future by taking actions in the present, which are selected from an appropriate policy. By construction, the expected free energy strikes the Bayes-optimal balance between goal-directed and information-seeking behavior under some prior preferences. It is this particular aspect of active inference that subsumes Bayesian decision theory and optimal Bayesian design principles as limiting cases (Da Costa, Parr, et al., 2020; K. Friston et al., 2020). To make the connections between these clearer, we highlight the different ways of unpacking expected free energy (Eq.3):

$$
\begin{aligned}
G(\pi,\tau) &= \mathrm{E}_{\tilde{Q}}[\log Q(s_\tau \mid \pi) - \log P(o_\tau, s_\tau)] \\
&= \underbrace{D_{KL}[Q(s_\tau \mid \pi) \,\|\, P(s_\tau)]}_{Risk} + \underbrace{\mathrm{E}_{Q(s_\tau|\pi)}\left[\mathrm{H}[P(o_\tau \mid s_\tau)]\right]}_{Ambiguity} \\
&\geq -\underbrace{\mathrm{E}_{Q(o_\tau|\pi)}\left[D_{KL}[Q(s_\tau \mid o_\tau, \pi) \,\|\, Q(s_\tau \mid \pi)]\right]}_{Intrinsic\ value} - \underbrace{\mathrm{E}_{Q(o_\tau|\pi)}[\log P(o_\tau)]}_{Extrinsic\ value}
\end{aligned} \tag{5}
$$

The second equality presents expected free energy as a balance between risk and ambiguity (Da Costa, Parr, et al., 2020). Here, risk is the difference between predicted and prior beliefs about future latent states. Thus, policies will be more probable if they lead to states that align with prior preferences, i.e., risk minimizing policies. This part of expected free energy underwrites KL control (Todorov, 2008; van den Broek et al., 2010). Conversely, ambiguity—the expectation of the conditional entropy—is the uncertainty about future outcome given beliefs about future latent states. This can also be interpreted as the expected inaccuracy of future predictions. Low ambiguity means that observed outcomes are uniquely informative about the latent states. Therefore, ambiguity minimization corresponds to selecting policies that lead to unambiguous and salient outcomes.

Equivalently, the expected free energy bounds the difference between intrinsic value (about states) and extrinsic value (Eq.5; third equality). These terms capture the imperative to maximize intrinsic value (i.e., information gain), from interactions with the environment, about the latent states, whilst maximizing extrinsic value (i.e., expected value), in relation to prior beliefs. Extrinsic value is encoded by log prior preferences over outcomes and is analogous to a cost or reward function in reinforcement learning (A. Barto, Mirolli, & Baldassarre, 2013; Cullen, Davey, Friston, & Moran, 2018; Sajid et al., 2021; Sutton & Barto, 1998). This encourages exploitative behavior such that policies are more likely if they help the agent solicit preferred outcomes. Maximizing intrinsic value promotes curious, information-seeking behavior as the agent seeks out salient latent states to minimize its uncertainty about the environment. This term underwrites artificial curiosity (J Schmidhuber, 1991; Schmidhuber, 2006), and for particular generative model parameterizations, it can be further decomposed to include expected information gain about model parameters (i.e., novelty) (Schwartenbeck et al., 2019). This particular decomposition of expected free energy speaks to the two aspects of goal-directed, information-seeking behavior.

## 3.1 Optimal Bayesian design and expected free energy

Optimal Bayesian design is concerned with maximizing information gain, i.e., resolving uncertainty about the latent states that generated particular outcomes (Chaloner & Verdinelli, 1995; Lindley, 1956; Pukelsheim, 2006; Stone, 1959). Therefore, maximizing intrinsic value or expected information gain—pertaining latent states or model parameters—has a direct correspondence to optimal Bayesian design principles. That is, if we were to remove prior preferences about outcomes—the extrinsic value component of expected free energy (Eq.5; third equality)—active inference reduces to optimal Bayesian design:

$$
\begin{aligned}
&\mathrm{E}_{Q(o_\tau|\pi)}[\log P(o_\tau)] = 0 \Rightarrow \\
&\qquad G(\pi,\tau) \geq -\mathrm{E}_{Q(o_\tau|\pi)}\left[D_{KL}[Q(s_\tau \mid o_\tau,\pi) \,\|\, Q(s_\tau \mid \pi)]\right] \\
&\qquad\qquad = -D_{KL}\left[Q(s_\tau,o_\tau \mid \pi) \,\|\, Q(s_\tau \mid \pi)Q(o_\tau \mid \pi)\right] \\
&\qquad\qquad = -I(\pi)
\end{aligned}
\tag{6}
$$

This is mathematically equivalent to expected Bayesian surprise, and mutual information that underwrites salience in visual search (Itti & Baldi, 2009; Sun, Gomez, & Schmidhuber, 2011) and the deployment of our visual epithelia with saccadic eye movements (H. Barlow, 1961; H. B. Barlow, 1974; Linsker, 1990; Optican & Richmond, 1987).

Within optimal Bayesian design principles, policies are regarded as a consequence of observed outcomes. This enables policy selection to maximize the evidence for a generative model, under which policies maximize information gain. To further elucidate this connection to active inference, we first establish a free energy functional of the predictive distribution that furnishes an upper bound on the expected log evidence of future outcomes. From this, posterior over policies can be evaluated:

$$
\begin{aligned}
&Q(\pi) = \arg\min_Q \mathrm{F}_\tau[Q(s_\tau,\pi)] \\
&\quad \mathrm{F}_\tau = \mathrm{E}_Q[\log Q(s_\tau,\pi) - \log P(o_\tau,s_\tau)] \\
&\qquad = \mathrm{E}_{Q(\pi)}[\mathrm{G}(\pi,\tau) + \log Q(\pi)] \\
&\qquad = \mathrm{E}_{Q(\pi)}[\mathrm{G}(\pi,\tau)] - \mathrm{H}[Q(\pi)] \\
&\qquad \Rightarrow -\log Q(\pi) = \mathrm{G}(\pi,\tau)
\end{aligned}
\tag{7}
$$

This renders the posterior surprisal of a policy its expected free energy $\mathrm{G}(\pi,\tau)$ (Da Costa, Parr, et al., 2020). Consequently, under prior beliefs that the log probability of a policy corresponds to information gain, the expected free energy of a policy provides an upper bound on the expected log evidence (i.e., marginal likelihood) over outcomes and policies. This can be shown by rearranging the expected free energy summands and introducing an expected evidence bound:

$$
\begin{aligned}
\mathrm{G}(\pi,\tau) &= \mathrm{E}_{Q(o_\tau,s_\tau|\pi)}[\log Q(s_\tau \mid \pi) - \log P(o_\tau, s_\tau)] \\
&= -\underbrace{\mathrm{E}_{Q(o_\tau|\pi)}[D_{KL}[Q(s_\tau \mid o_\tau,\pi) \,||\, Q(s_\tau \mid \pi)]]}_{\textit{Expected information gain}} - \underbrace{\mathrm{E}_{Q(o_\tau)}[\log P(o_\tau)]}_{\textit{Expected log evidence}} \\
&\quad + \underbrace{\mathrm{E}_{Q(o_\tau|\pi)}[D_{KL}[Q(s_\tau \mid o_\tau,\pi) \,||\, P(s_\tau \mid o_\tau)]]}_{\textit{Expected evidence bound}} \\
&\geq -\underbrace{\mathrm{E}_{Q(o_\tau|\pi)}[\log P(\pi \mid o_\tau)]}_{\textit{Expected information gain}} - \underbrace{\mathrm{E}_{Q(o_\tau)}[\log P(o_\tau)]}_{\textit{Expected log evidence}} \\
&= -\underbrace{\mathrm{E}_{Q}[\log P(o_\tau,\pi)]}_{\textit{Expected marginal likelihood}}
\end{aligned}
\tag{8}
$$

where:

$$
\log P(\pi \mid o_\tau) = \underbrace{D_{KL}[Q(s_\tau \mid o_\tau,\pi) \,||\, Q(s_\tau \mid \pi)]}_{\textit{Information gain}}
$$

In short, policies depend on the final outcomes, and are more likely when outcomes reduce uncertainty about latent states. That is, the expected free energy of a policy is an upper bound on the expected log evidence for a model that generates outcomes, and Bayes optimal policies from those outcomes. Consequently, when the expected evidence bound is minimized, the predictive posterior becomes the posterior under the generative model.

## 3.2 Bayesian decision theory and expected free energy

Bayesian decision theory—i.e., Bayesian formulations of maximizing expected utility under uncertainty—is predicated on the optimization of some expected cost or utility function (Berger, 2011; Harsanyi, 1978; Savage, 1972). Therefore, maximizing the extrinsic value (or expected value) in relation to some prior beliefs under active inference, has a direct correspondence to Bayesian decision theory. This can be derived in two ways. First, by removing the expected information gain, from the expected free energy objective (Eq.5; third equality)—active inference reduces to Bayesian decision theory:

$$
\begin{aligned}
&\mathrm{E}_{Q(o_\tau|\pi)}\left[D_{KL}[Q(s_\tau \mid o_\tau,\pi) \,||\, Q(s_\tau \mid \pi)]\right] = 0 \Rightarrow \\
&\qquad\qquad G(\pi,\tau) \geq -\mathrm{E}_{Q(o_\tau|\pi)}[\log P(o_\tau)]
\end{aligned}
\tag{9}
$$

Now, we are left with extrinsic value or expected utility in economics (Von Neumann & Morgenstern, 1944). This is mathematically equivalent to the (reward) objectives employed in reinforcement learning and behavioral psychology (Da Costa, Sajid, Parr, Friston, & Smith, 2020; Sajid et al., 2021; Sutton & Barto, 1998). Notice, that here the expected utility is formulated in terms of outcomes.

Conversely, expected utility can also be derived in terms of latent states by removing ambiguity and the posterior entropy from the expected free energy objective (Eq.5):

$$\begin{aligned} & \mathrm{E}_{Q(s_\tau, o_\tau \mid \pi)}[\log Q(s_\tau \mid \pi) \underbrace{- \log P(o_\tau \mid s_\tau)}_{Ambiguity}] = 0 \Rightarrow \\ & \qquad G(\pi, \tau) = -\mathrm{E}_{Q(s_\tau \mid \pi)}[\log P(s_\tau)] \end{aligned} \tag{10}$$

This reduced formulation entails the maximization of some utility function, expected under predictive posterior beliefs about the consequences of action, and:

$$\mathrm{E}_Q[\log P(s_\tau)] \le \mathrm{E}_Q[\log P(o_\tau)] \tag{11}$$

This equivalence shows that maximizing expected utility under uncertainty, can be defined as the optimization of some expected cost, either in terms of latent states or observed outcomes. By definition, both formulations should yield comparable behavior. However, by only maximizing the expected value, Bayesian decision theory is only optimal when specific conditions (those before the implications in Equations 9 or 10) are met. Specifically, from an active inference perspective, Bayes optimality is a direct consequence of making decisions that maximize the expected free energy, not expected utility.

## 4.0 Simulations

In the previous section, we saw that both Bayesian decision theory and optimal Bayesian design are special cases of minimizing expected free energy. Specifically, Bayesian decision theory is predicated on optimizing some expected cost, and optimal Bayesian design on maximizing information gain. In active inference, we directly optimize both these distinct objectives:

*Active inference = Bayes decision theory + optimal Bayesian design* (12)

To illustrate how behavioral differences arise under these separable imperatives, we consider inference using simulations of foraging in a maze, where the agent selects the next action by optimizing the following: 1) expected utility, 2) expected information gain or 3) expected free energy. For ease of exposition, we have purposely chosen a simple paradigm. More complex active inference simulations can be seen in narrative construction and reading (K. J. Friston et al., 2020; Karl J. Friston, Rosch, Parr, Price, & Bowman, 2018), saccadic searches, and scene construction (Mirza, Adams, Mathys, & Friston, 2016; Parr, 2019), 3D mazes (Fountas et al., 2020), etc.

First, we describe the maze foraging environment, and the accompanying generative model from (K. Friston et al., 2017; Karl J. Friston et al., 2015). We then present the simulation results of how behavior differs when removing different constituents of expected free energy.

## 4.1 Maze foraging environment

In this setup, a mouse starts at the center of the T-maze: it can either move directly to the right or left arms—that contains some cheese—or to the lower arm that contains cues about whether the cheese is in the upper right or left arm. The agent can only move twice and upon entering the upper right or left arms cannot leave i.e., these are absorbing states. Thus, the optimal behavior is to first go to the lower arm to find the reward location and then retrieve the reward. If the agent follows this path, it is given a performance score of $+5$ from the environment, if it goes directly to the correct cheese location it receives a score of +10 but failure to find the correct cheese location results in -10. These scores just allow us to record the agent's performance and play no part in the agent's decision-making process: see Figure 1. Consequently, what the agent considers to be optimal (defined by its prior preferences over outcomes), may or may not coincide with a high 'performance' score.

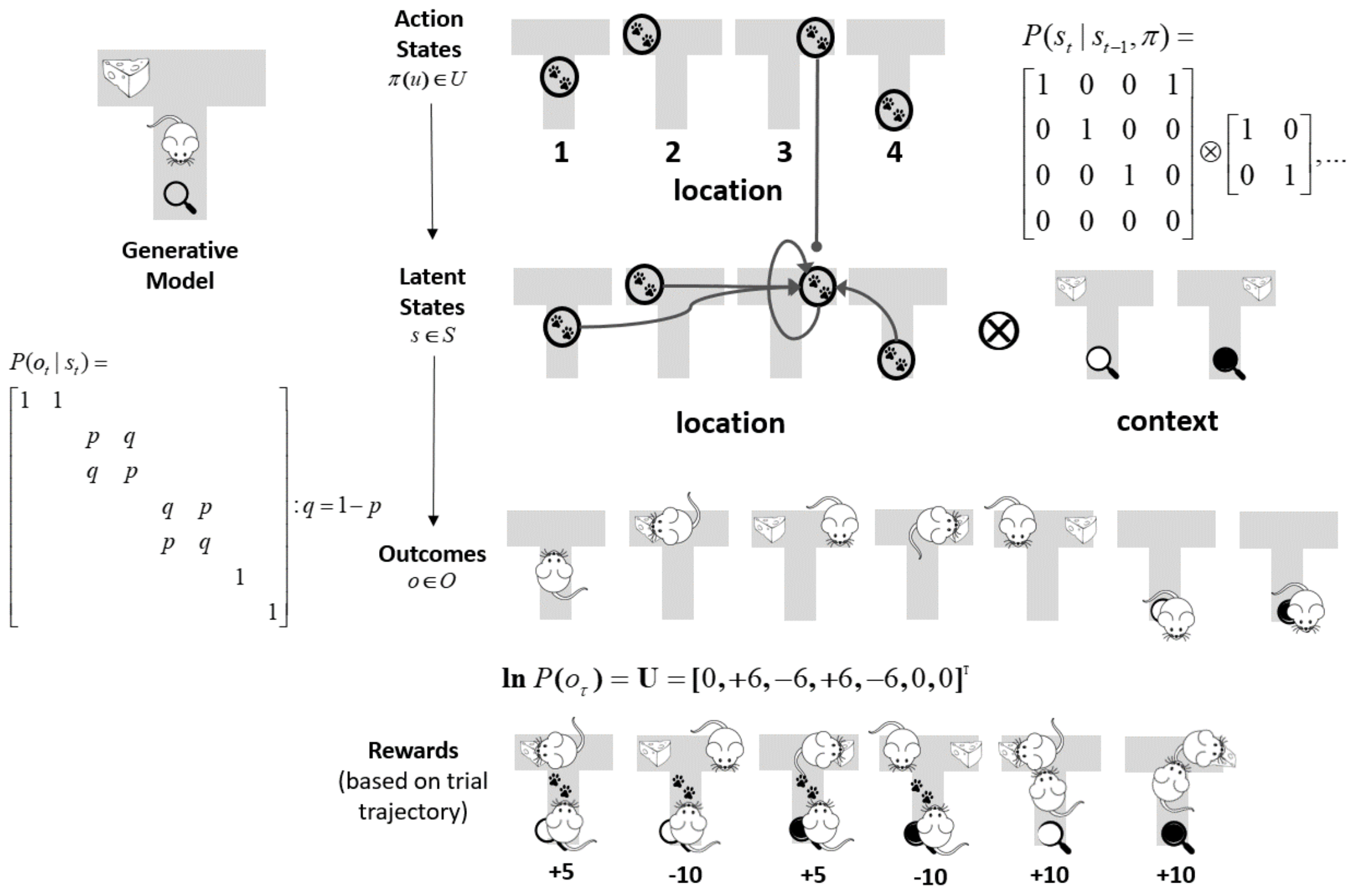

**Figure 1.** A generative model of the T-maze task- adapted from (K. Friston et al., 2017; Karl J. Friston et al., 2015). The model contains four action states that encode movement to one of the four locations: center, lower arm, upper right, and left arm. These states control the ability to transition between the latent states that have a Kronecker tensor product ($\otimes$) form with two factors: location (one of the four) and context (one of the two). These correspond to the location of the cheese (reward) and associated cues (white or black). From each of the eight latent states—an observable outcome is generated, and the first two latent states generate the same outcome that simply tells the agent that it is at the center. A few selected transitions have been shown (via arrow on the figure), indicating that action states attract the agent to different locations, where outcomes are sampled. Categorical parameters that define the generative model—**A** (latent states to outcomes) & **B** (state transitions)—have been explicitly included. Additionally, $\ln P(o)$ corresponds to prior preferences. A preference of 6 or -6 is allocated to the correct or incorrect cheese location, respectively (with preferences of zero otherwise). These values relate to the unnormalized log probability. While the numerical differences between them are preserved, the absolute values may change on normalization.

## 4.2 Generative model of maze foraging

We define the generative model as follows: four control states that correspond to visiting the four locations (the center and three arms—we assume each control state takes the agent to the associated location), eight latent states (four locations factorized by two contexts) and seven possible outcomes. The outcomes correspond to being in the center, plus the (two) outcomes at each of the (three) arms that are determined by the context, i.e., whether the right or the left arm has the cheese. We define the likelihood, $P(o_\tau \mid s_\tau)$ such that the ambiguous cue is at the center (first) location and a definitive cue at the lower (fourth) location (refer to Figure 1). The remaining locations provide a reward with probability $p = 98\%$ determined by the context. The transition probabilities, $P(s_\tau \mid s_{\tau-1}, \pi)$, encode how the mouse might move, except for the second and third locations, which are absorbing latent states that it cannot leave. We define the mouse as having precise beliefs about all contingencies, except the current context. Each context is equally probable. Additionally, we remove the agent's capacity to learn these contingencies based on interactions with the environment. In other words, historic trials have no bearing on the current trial. The utility of the outcomes is: 6 and -6 for identifying the correct and incorrect cheese location, respectively (and zero otherwise). Having specified the state-space and contingencies, we can perform gradient descent on the free energy functionals to simulate behavior. Relative prior beliefs about the initial state were initialized to 128 for the central location for each context and zero otherwise.

## 4.3 Simulations of maze foraging

To illustrate the relationship between planning and behavior under active inference, optimal Bayesian design, and Bayesian decision theory, we simulated three agents performing the maze foraging task. The active inference agent optimized expected free energy, optimal Bayesian design optimized expected information gain, and Bayesian decision theory optimized expected utility. Each simulation comprised 50 trials, with shifting context, i.e., occasionally moving the cheese from the right to the left arm. The context, indicated by the (black or white) cue in the lower arm, was white until trial 9, black from trial 10 to 12, white again from trial 13 to 29, and black again at trial 30. After trial 30, it remained white until the end of the simulation. These switches allowed us to evaluate behavioral shifts from information-seeking to goal-directed policies. Everything else was kept constant, including the initial conditions. Belief updating and behavior were simulated using the variational message passing scheme implemented in SPM::spm_MDP_VB_X.m (http://www.fil.ion.ucl.ac.uk/spm/).

During the first trial, we observe marked differences in behavior (Figure 2; action selection). Unsurprisingly, the active inference agent can trade-off between information-seeking (i.e., go to the cue at epoch 2) and extrinsic value (i.e., go to the left arm to collect the cheese at epoch 3). Conversely, the optimal Bayesian design agent exhibits purely information maximizing behavior, i.e., go to the cue at epoch 2, and then remaining there till the end of the trial. This follows directly from resolving its uncertainty about the trial context, i.e., there is no further information to be gained: anecdotally, the agent gets bored and does nothing after exploring its environment. This behavior can be regarded as Bayes-optimal; when given uniform prior preferences, purely epistemic, information-seeking behavior is the only appropriate way to forage the maze. The Bayesian decision theory agent follows a different strategy: it remains in the central location till epoch 2 after which it goes to the right arm (i.e., the arm without the cheese). This is expected since the agent does not care about the information inferred and is indifferent about going to the cue location or remaining at the central location.

Predictably, these behavioral differences influence both the position and context latent state estimations (Figure 2; position and context). As the agents select different actions to navigate the maze, their location estimates differ along with their beliefs about the current context. The active inference agent estimates its location to be in the left arm at epoch 3 and can accurately infer that the current context is white. On the other hand, the Bayesian decision theory agent infers itself to be at the right arm at epoch 3. This inference is underwritten by the fact that it collects a reward, which also has epistemic value. Whilst, the optimal Bayesian design agent believes itself, correctly, to be in the lower arm at epoch 3, having resolved its uncertainty about the current states of affairs.

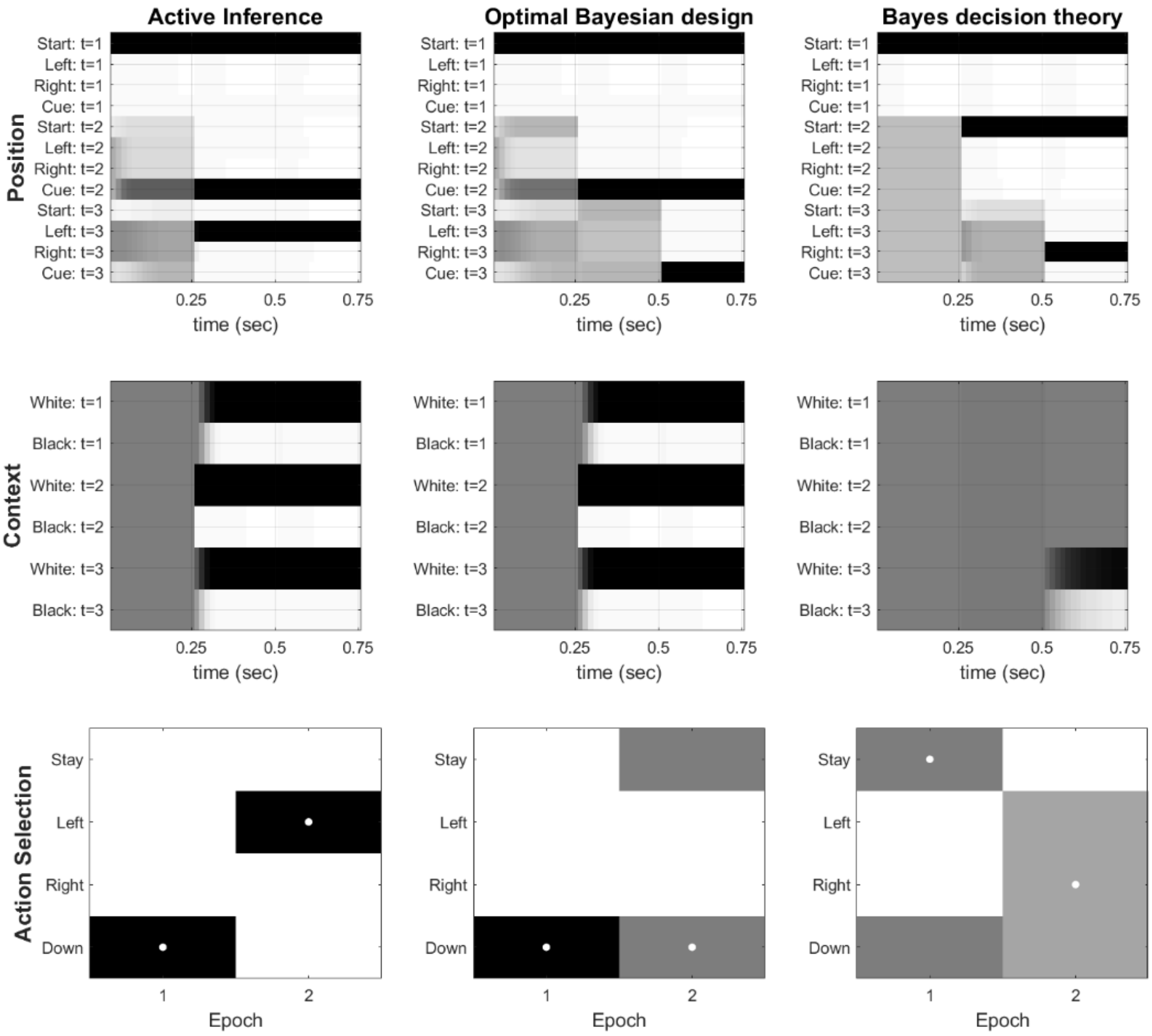

**Figure 2.** Belief updating and action selection. The graphics present the belief updates and action selection for the first simulated trial. The first two rows report belief updating over three epochs of a single trial, for the latent states position and context. The bottom row presents the actions selected during the same trial. Each column represents a simulated agent: active inference, optimal Bayesian design, and Bayesian decision theory. Here, white is an expected probability of zero, black of one, and grey indicates gradations between these. For the belief updating figures, the *x*-axis represents time in seconds (divided into 3 epochs), and the *y*-axis represents posterior expectations about each of the associated states at different epochs (in the past or future). For example, for the position latent state (first row), there are 4 states (start, left, right, and cue), and a total of 4 x 3 (states times epochs) posterior expectations—similarly, for the context latent state (second row), there are 2 levels, and a total of 2 x 3 expectations. For example, the first four rows for the position belief updates, correspond to expectations about the agent's position, in terms of the four alternatives for the first epoch. The second four rows are the equivalent expectations for the second epoch, and so on. This means that at the beginning of the trial the second four rows report beliefs about the future: namely, the next epoch. However, later in time, these beliefs refer to the past, i.e., beliefs currently held about the first epoch. Note that most beliefs persist through time (along the *x*-axis), endowing the agents with a form of working memory that is both predictive and postdictive. In the action selection panels, the *x*-axis represents the 2 epochs, and the y axis represents the four allowable actions with their posterior expectations at each epoch: stay center, go down, turn left, or right. In each graphic, the action with the circle marker indicates the selected action.

Subsequent trials reveal comparable behavior (Figure 3; policy selection). The active inference agent pursues an epistemic policy throughout the 50 trials, i.e., first, go to the cue and then the cheese location. This is entirely appropriate since at the start of each trial the agent does not know where the cheese is located and the only way to resolve its uncertainty is to go to the cue location. This is the only agent that consistently selects information-seeking, goal-directed policies, as reflected by the high accumulated score (Figure 3; score).

In contrast, the Bayesian decision theory agent (i.e., maximizing expected utility), entertains all sorts of policies. However, the pattern is consistent. First, it chooses between remaining at the central location or going to the cue location with equal probability. If the agent goes to the cue location, then the only plausible action chosen is to go to the left arm. Instead, if it decides to remain at the central location, it can decide to go to the left, right, or lower arm. This is because it is unable to resolve uncertainty about the current context.

The optimal Bayesian design agent selects information-seeking policies, but, unsurprisingly, does not prefer to end the trial by going to either left or right arm. Instead, it chooses to remain at the cue location or go to the central location again. This intrinsically motivated behavior is reflected in the low accumulated score (Figure 3; score). None of the agents selected the exploitative policies i.e., go to either left or right arm at epoch 2 and stay there. This is due to the uncertainty about which latent states generated the particular observations, i.e., an inability to *learn* the context, which would be appropriate if the cheese were located randomly for every trial.

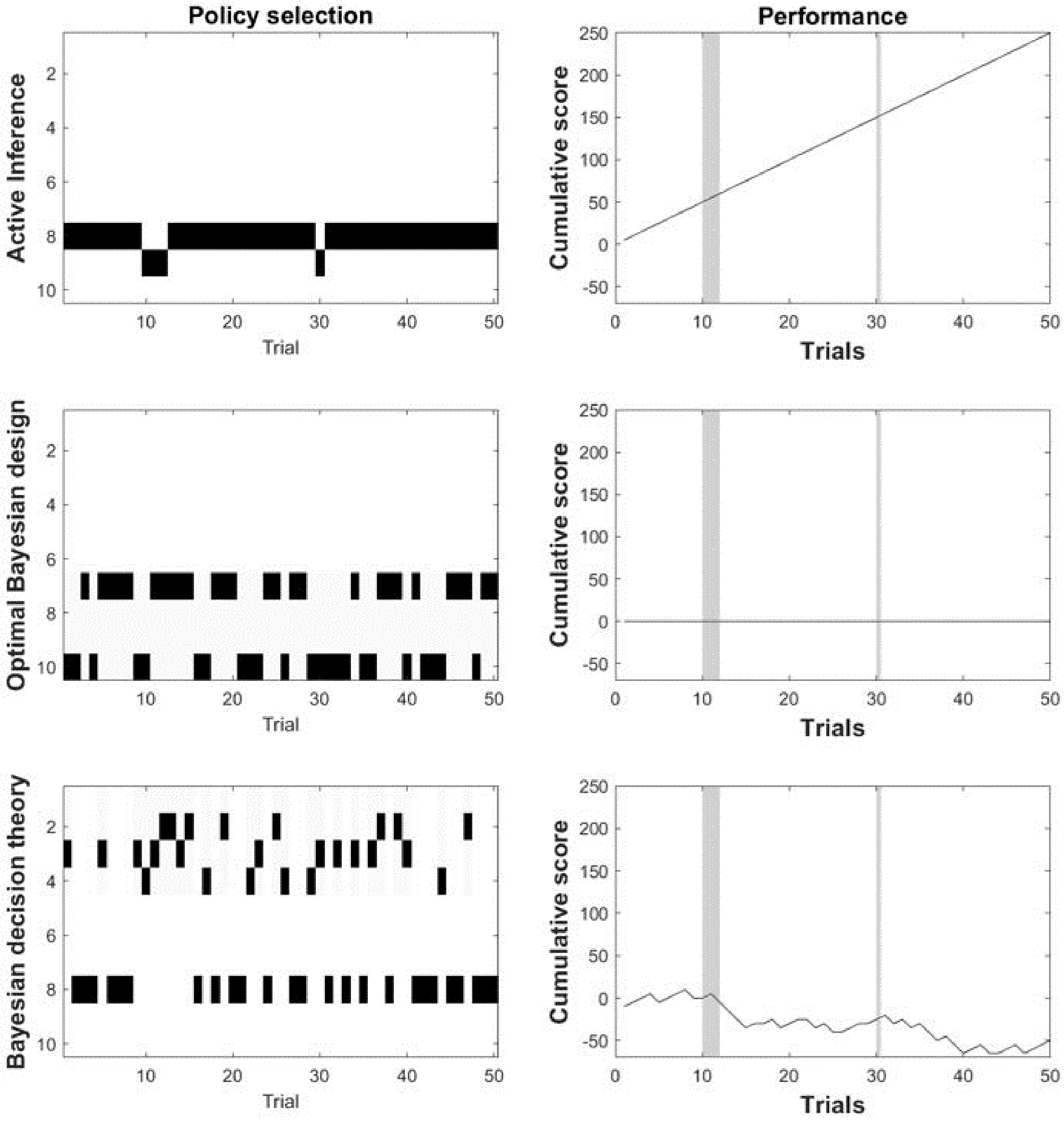

**Figure 3.** Policy selection and performance. The graphics report the selected policies, accumulated score, and optimization trajectory during the 50 simulated trials. The first column reports the conditional expectations over policies for each agent (the corresponding rows), the second column reports the accumulated score evaluated using the scoring presented in Figure 1. For the policy selection panels on the left, the *x*-axis represents trials, and the *y*-axis represents the final posterior expectations about the ten policies across the 50 trials. The ten policies are: 1—continue to stay at the starting location, 2—stay at the center and then to go the left arm, 3—stay at the center and then to go the right arm, 4—stay at the center and then to go the lower arm, 5—go to the left arm and stay, 6—go the right arm and stay; 7—go to the cue and then back to the center, 8—go to the cue and then left arm, 9—go to the cue and then right arm and 10—go to the cue and stay. Here, white is an expected

probability of zero, black of one, and grey indicates gradations between these. For the score panels on the right, the *x*-axis represents trials, and the *y*-axis represents the cumulative score. The grey rectangles denote a black context, while their absence reflects a white context.

## 5.0 Concluding remarks

Active inference scores the consequences of action in terms of the expected free energy that can be decomposed into two parts. The first part corresponds to the information gain that underwrites optimal Bayesian design. Thus, by removing any prior preferences over observations from the expected free energy, we are left with optimal Bayesian design imperatives; namely, minimizing uncertainty about states of affairs. This can be operationalized as maximum entropy sampling (Mitchell, Sacks, & Ylvisaker, 1994; Sacks, Welch, Mitchell, & Wynn, 1989; Shewry & Wynn, 1987), augmented with ambiguity aversion. The second part encodes expected model evidence—or marginal likelihood—that can be associated with an expected utility in Bayesian decision theory. Consequently, active inference subsumes Bayesian decision theory and optimal Bayesian design. Notably, expected free energy can be further decomposed to derive other special cases, including maximum entropy principle (Jaynes, 1957), Occam's principle, etc. (K. Friston et al., 2020). This speaks to the potentially ubiquitous nature of active inference as realizing a large class of sentient systems that self-organize themselves to some non-equilibrium steady-state (K. Friston, 2019; Parr et al., 2020). This means that a certain class of systems at steady-state will appear to operate Bayes-optimally—both in terms of optimal Bayesian design (i.e., information-seeking or explorative behavior) and Bayesian decision theory (goal-directed or exploitative behavior) (Da Costa, Parr, et al., 2020).

This way of characterizing active inference, and the expected free energy, speaks to a natural emergence of epistemic behavior. Information-seeking means that the agent seeks out states that afford observations, which minimize uncertainty about latent states. The maze foraging simulations highlight this property of active inference. Due to uncertainty about the context, and causes of observed outcomes, the agent engages in exploratory behavior. This curiosity is manifest by choosing policies that first lead to the lower arm to disclose the cue, which allowed the agent to determine the cheese location. In other words, the agent first engages in exploratory, epistemic foraging and then exploits its beliefs about states of affairs. Notice that even though the cheese is left in the same place for several trials, the agent sticks to its preferred policy, since we do not equip it with the ability to learn that the context can persist over several trials. In the absence of uncertainty, the agent would choose exploitative policies, i.e., forego its epistemic foraging (K. Friston et al., 2017; Schwartenbeck et al., 2019).

Accordingly, implementations of active inference depend entirely on the generative model, e.g., the inclusion of particular priors, different hierarchical levels, etc. (K. J. Friston, Parr, & de Vries, 2017). Specifically, if one were to remove particular priors over the generative model, different types of behavior would emerge. For example, when no preferred state has been specified, active inference reduces to optimal Bayesian design principles. In contrast, when the distribution over preferred states is a point mass, active inference reduces to expected utility theory (Da Costa, Sajid, et al., 2020). This leads to sub-optimal behavioral convergence, by constraining agent interaction to specific parts of the environment (Tschantz et al., 2020). In the same spirit, one could also complexify the generative model with additional parameters that could be inferred. This could introduce a shift in the balance between intrinsic and extrinsic value. For example, depending on the context, or the environment, prior preferences could be emphasized by increasing their precision (Da Costa, Parr, et al., 2020).

Conversely, in the reinforcement learning literature, different types of information-seeking behaviors have been engineered by supplementing objective functions or aspects of the implicit model. In contrast to active inference, these augmented reinforcement learning schemes try to maximize the agent's surprise (i.e., inability to predict the future). In temporal difference learning, this is often achieved by encouraging exploration through $\varepsilon$-greedy policies, where actions are random with probability $1-\varepsilon$ (i.e., random exploration). This reduces the probability of policy learning falling into local minima (Sutton & Barto, 1998). Other approaches focus on intrinsic motivation or curiosity rewards, and augment the objective function with entropy terms, thereby encouraging more entropic policies (A. G. Barto, 2013; Bellemare et al., 2016; Burda, Edwards, Storkey, & Klimov, 2018; Pathak, Agrawal, Efros, & Darrell, 2017; J Schmidhuber, 1991; Jürgen Schmidhuber, 1991). This encourages policies that increase information gain by exploring unpredictable states. Popular Bayesian reinforcement learning schemes—for inducing information-seeking behavior—include upper confidence bounds (UCB), optimistic Bayesian sampling and Thompson sampling, and variational inference approaches like VIME and variBAD (Auer, 2002; Houthooft et al., 2016; Russo, Van Roy, Kazerouni, Osband, & Wen, 2017; Schulz & Gershman, 2019; Zintgraf et al., 2019). These variational approaches are perhaps closest to active inference in the way they are set-up. However, differences arise in how actions are selected e.g., VIM selects future actions by minimizing the entropy of beliefs about transition probabilities (Houthooft et al., 2016). The VIM scheme assumes an observable MDP and therefore ignores an important source of uncertainty, in relation to minimizing the expected free energy in active inference. Finally, note that reward maximizing schemes are a special case of active inference (Da Costa, Sajid, et al., 2020), when certain sources of uncertainty are ignored. In other words, in the absence of risk and ambiguity, the principle of least action—that underwrites active inference—reduces to the Bellman optimality principle (K. Friston et al., 2016).

In this chapter, we have discussed the natural emergence of information-seeking behavior under active inference. How this may be realized in sentient creatures remains an open, and interesting research area (Gottlieb, Oudeyer, Lopes, & Baranes, 2013). For example, humans employ a combination of epistemic and extrinsic value policies when engaging with their environment, but their information-seeking behavior can be modified by changes in temporal resolution (Vasconcelos, Monteiro, & Kacelnik, 2015), and/or temporal horizon (Wilson, Geana, White, Ludvig, & Cohen, 2014). This suggests that future work should consider which structural or parametric aspects of generative models (Vértes & Sahani, 2018) best account for individual differences in information-seeking behavior.

**Acknowledgments**

NS is funded by Medical Research Council (MR/S502522/1). LD is supported by the Fonds National de la Recherche, Luxembourg (Project code: 13568875). KJF is funded by the Wellcome Trust (Ref: 088130/Z/09/Z).